\documentclass[sigconf]{acmart} 

\AtBeginDocument{%
  \providecommand\BibTeX{{%
    \normalfont B\kern-0.5em{\scshape i\kern-0.25em b}\kern-0.8em\TeX}}}

\copyrightyear{2021}
\acmYear{2021}
\setcopyright{acmcopyright}\acmConference[MM '21]{Proceedings of the 29th ACM
International Conference on Multimedia}{October 20--24, 2021}{Virtual Event, China}
\acmBooktitle{Proceedings of the 29th ACM International Conference on Multimedia
(MM '21), October 20--24, 2021, Virtual Event, China}
\acmPrice{15.00}
\acmDOI{10.1145/3474085.3475551}
\acmISBN{978-1-4503-8651-7/21/10}

\acmSubmissionID{2056}

\usepackage{subfigure}
\usepackage{stfloats}
\usepackage{amsmath, lipsum}
\usepackage{cuted}
\usepackage{amsmath}
\usepackage{indentfirst}
\usepackage[noend]{algpseudocode}
\usepackage{algorithmicx,algorithm}

\usepackage{setspace}
\usepackage{balance}
\usepackage{fancyhdr}

\begin{document}

\title{Dense Semantic Contrast for \\ Self-Supervised Visual Representation Learning}


\newcommand\blfootnote[1]{
\begingroup
\renewcommand\thefootnote{}\footnote{#1}
\addtocounter{footnote}{-1}
\endgroup
}
\settopmatter{printacmref=false}

\author{Xiaoni Li$^{1,2}$, Yu Zhou$^{1,2*}$, Yifei Zhang$^{1,2}$, Aoting Zhang$^1$}
\author{Wei Wang$^{1,2}$, Ning Jiang$^3$, Haiying Wu$^3$, Weiping Wang$^1$}
\affiliation{
\institution{$^1$Institute of Information Engineering, Chinese Academy of Sciences, Beijing, China\\$^2$School of Cyber Security, University of Chinese Academy of Sciences, Beijing, China\\$^3$Mashang Consumer Finance Co., Ltd., Beijing, China}
\city{}
  \country{}}
\email{{ lixiaoni, zhouyu, zhangyifei0115, wangwei3456, wangweiping}@iie.ac.cn}
\email{aotingzhang@126.com, {ning.jiang02, haiying.wu02}@msxf.com}

\renewcommand{\shortauthors}{Xiaoni and Yu, et al.}
\fancyhf{}
\begin{abstract}
\noindent Self-supervised representation learning for visual pre-training has achieved remarkable success with sample (instance or pixel) discrimination and semantics discovery of instance, whereas there still exists a non-negligible gap between pre-trained model and downstream dense prediction tasks. Concretely, these downstream tasks require more accurate representation, in other words, the pixels from the same object must belong to a shared semantic category, which is lacking in the previous methods. In this work, we present Dense Semantic Contrast (DSC) for modeling semantic category decision boundaries at a dense level to meet the requirement of these tasks. Furthermore, we propose a dense cross-image semantic contrastive learning framework for multi-granularity representation learning. Specially, we explicitly explore the semantic structure of the dataset by mining relations among pixels from different perspectives. For intra-image relation modeling, we discover pixel neighbors from multiple views. And for inter-image relations, we enforce pixel representation from the same semantic class to be more similar than the representation from different classes in one mini-batch. Experimental results show that our DSC model outperforms state-of-the-art methods when transferring to downstream dense prediction tasks, including object detection, semantic segmentation, and instance segmentation. Code will be made available.
 
\end{abstract}






\maketitle
\vspace{15px}
\begin{small}
\begin{spacing}
1
\noindent\textbf{ACM Reference Format:}

\noindent Xiaoni Li, Yu Zhou, Yifei Zhang, Aoting Zhang, Wei Wang, Ning Jiang, Haiying Wu, and Weiping Wang. 2021. Dense Semantic Contrast for Self-Supervised Visual Representation Learning. In \textit{Proceedings of the 29th ACM International Conference on Multimedia (MM ’21), October 20-24, 2021, Virtual Event, China.} ACM, New York, NY, USA, 9 pages. \\ https://doi.org/10.1145/3474085.3475551
\end{spacing}
\end{small}

\blfootnote{*Yu Zhou is the corresponding author.}

\section{Introduction}
\begin{figure}
  \centering
  \includegraphics[width=0.9\linewidth]{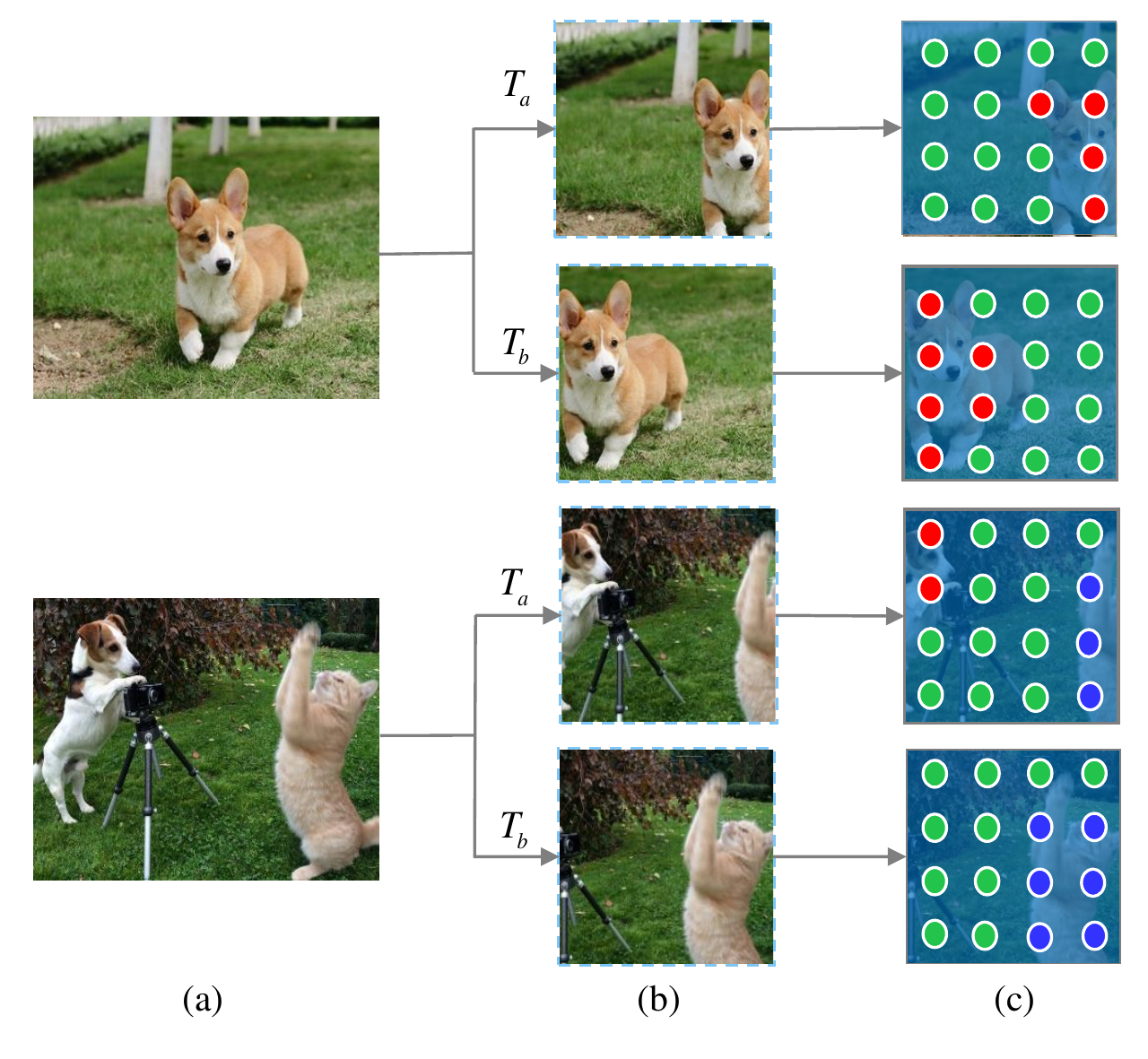}
  \caption{Illustration of pixels with same semantics under intra- and inter-image. The pixels from a same object share a common semantic category, which should be drawn closer.}
  \label{motivation}
\end{figure}

Despite that self-supervised pre-training \cite{ldz1,ldz2,ldz3,lw1} has achieved breakthrough performance with large-scale datasets ($e.g.$, ImageNet \cite{DBLP:conf/cvpr/DengDSLL009}), the gap between the pre-trained model and the downstream dense prediction tasks (such as object detection \cite{DBLP:journals/ijcv/EveringhamGWWZ10,DBLP:conf/eccv/LinMBHPRDZ14,ydb1,ydb2,jn,qxg1,qxg2,qz1,qz2,cyd1,cyd2} and segmentation \cite{DBLP:conf/cvpr/CordtsORREBFRS16}) is still unnegligible. The object detection task aims to predict categories and bounding boxes for all the objects of interest in the image, while the objective of the segmentation task is to assign a category for each pixel. All these tasks need denser and semantic representation for more precise prediction. However, the previous instance-level self-supervised learning (SSL) methods \cite{DBLP:conf/icml/ChenK0H20,DBLP:conf/cvpr/He0WXG20,DBLP:journals/corr/abs-2003-04297,DBLP:conf/cvpr/WuXYL18,DBLP:conf/nips/GrillSATRBDPGAP20,DBLP:journals/corr/abs-2011-10566,DBLP:conf/eccv/CaronBJD18,DBLP:conf/iclr/AsanoRV20a,DBLP:conf/eccv/NiuZWL20,DBLP:conf/cvpr/HuangGZ20,DBLP:conf/cvpr/ZhanX0OL20,DBLP:conf/iccv/JiVH19,DBLP:conf/nips/CaronMMGBJ20,DBLP:journals/corr/abs-2103-02662,zyf1} for pre-training obtain only global feature representation, which is more fit for global classification. Additionally, recently developed pixel-level methods \cite{DBLP:conf/nips/PinheiroABGC20,DBLP:journals/corr/abs-2011-10043,DBLP:journals/corr/abs-2011-09157} are limited in a finite level due to the lack of semantic category decision boundary modeling.

The representation learning methods based on instance discrimination have recently achieved state-of-the-art performance by attracting positive samples while repelling negative samples. IR~\cite{DBLP:conf/cvpr/WuXYL18} proves that non-parametric instance-level classification can capture visual similarity. After that, some view-invariant approaches such as MoCo v1\&v2~\cite{DBLP:conf/cvpr/He0WXG20,DBLP:journals/corr/abs-2003-04297} and SimCLR~\cite{DBLP:conf/icml/ChenK0H20} propose that good representation can be learned by treating its augmented version as positive samples. 
Nowadays, BYOL~\cite{DBLP:conf/nips/GrillSATRBDPGAP20} and SimSiam~\cite{DBLP:journals/corr/abs-2011-10566} prove that positive pairs are sufficient for learning good feature representation without  negative pairs. However, these instance discrimination methods neglect the relations among different instances. To supply actual semantic category information in the dataset, some works~\cite{DBLP:conf/eccv/CaronBJD18,DBLP:conf/aaai/Huang0GZ20} are devoted to modeling semantic structures to reach supervised learning's performance. DeepCluster~\cite{DBLP:conf/eccv/CaronBJD18} and AND \cite{DBLP:conf/aaai/Huang0GZ20} are two typical semantic information exploring works implemented by clustering and discovering $k$ nearest neighbors. Additional works~\cite{DBLP:conf/cvpr/ZhanX0OL20,DBLP:conf/nips/CaronMMGBJ20,DBLP:conf/aaai/Huang0GZ20,DBLP:conf/iccv/ZhuangZY19} begin to explore semantic information to learn more discriminative feature representations subsequently. 
Nevertheless, both instance discrimination and semantics-discovery methods focus on global feature representation of images, and are only suitable for object-centric datasets such as ImageNet~\cite{DBLP:conf/cvpr/DengDSLL009}, which only contains one main object in each image. As shown in Figure~\ref{motivation}, instance discrimination methods treat each image as an individual class, ignoring the relation of the two images that both of them contain the same object ($i.e.$, dog). While semantics-discovery methods can't distinguish these two images as a positive pair because the objects they contain are not the same (a dog $vs.$ a dog and a cat), which reduced their similarity. Therefore, these approaches are not universal for downstream dense prediction tasks and the pre-training in complex authentic scenario images. 

To explore more suited pre-training approaches for dense prediction tasks, \cite{DBLP:conf/nips/PinheiroABGC20,DBLP:journals/corr/abs-2011-10043,DBLP:journals/corr/abs-2011-09157} conduct contrastive learning from a denser perspective with the notion of pixel discrimination. They treat each pixel as a single class and learn the discriminative representation for pixels. Although their specific design narrows the gap between the pre-trained model and the downstream dense prediction tasks evidently, they lack pixel-level semantic category discriminative capability since any non-linear intra-class variations in each pixel is not modeled. Hence, it is limited to low- and mid-level visual understandings on an individual pixel level. Take Figure~\ref{motivation} as an example again, the pixels in the region of the two dogs belong to the same category semantically (the red circles), which should be drawn closer to each other. Conversely, the pixels in the two dogs' regions are not the same as those in the region of the cat (the blue circles), which should be pushed away. However,  the previous pixel discrimination methods push all of those pixels away, lacking high-level visual understandings.

\begin{figure*}
  \centering
  \includegraphics[width=0.9\linewidth]{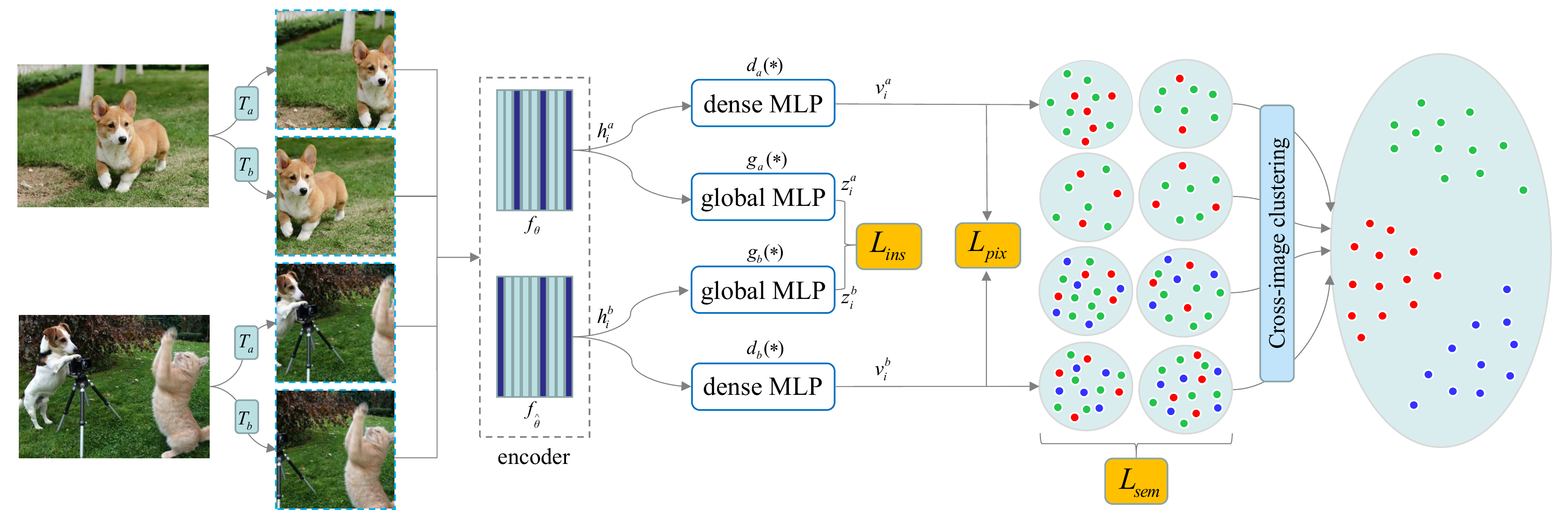}

  \caption{Architecture of the DSC framework for multi-granularity representation learning.}
  \label{pipeline}
\end{figure*}

In this work, we present the concept of Dense Semantic Contrast (DSC) for explicitly modeling semantic category decision boundaries at the pixel level, which establishes connections both instance-to-instance and pixel-to-pixel semantically. Besides, a dense cross-image semantic contrastive learning framework for multi-granularity representation learning is constructed to make up for semantics' defects compared with previous SSL pre-training methods. Specially, we first explore a neighbors-discovery method to enhance the correlation of the pixels within the image, which mines the neighbors from multiple views (Firgure~\ref{fig3}). Moreover, we design a dense semantic module for cross-image semantic relation modeling by adopting certain clustering methods shown in the right part of Figure~\ref{pipeline}. Here we focus on k-means (KM) and prototype mapping (PM) \cite{DBLP:conf/nips/CaronMMGBJ20} for simplicity. While other clustering methods can also be adopted, such as Power Iteration Clustering \cite{DBLP:conf/icml/LinC10} and Invariant Information Clustering \cite{DBLP:conf/iccv/JiVH19}. For the other granularities, we conduct instance and pixel discrimination performed standard contrastive learning. DSC is an end-to-end manner as it can consider multi-granularity contrastive representation learning simultaneously. To summarize, the major contributions of our work are three-fold:

1) For the first time, we reveal that the pixel discrimination task is short of semantic category decision boundary reasoning capability. The insufficiency of the ability leads that the transferred model can't assign the same category label for the pixels from one object accurately, resulting in the gap between the pre-trained models and the downstream dense prediction tasks. Consequently, we model the semantic decision boundary explicitly to narrow this gap.

2) We propose a dense cross-image semantic contrastive learning framework for multi-granularity representation learning. Unlike the previous SSL pre-training methods, the framework considers the semantic relations of both intra- and inter-image pixels. We learn the discriminative information in the instance, pixel, and pixel category granularity simultaneously to ensure the diversity of the intra-class features and the discrimination of the inter-class features at the pixel level.

3) We transfer the model pre-trained on ImageNet \cite{DBLP:conf/cvpr/DengDSLL009} and MS COCO \cite{DBLP:conf/eccv/LinMBHPRDZ14} to abundant downstream dense prediction tasks. All the experimental results show that DSC achieves superior or comparable performance with previous works \cite{DBLP:journals/corr/abs-2003-04297,DBLP:journals/corr/abs-2011-09157}, which once again proves the importance of semantic relation in self-supervised visual representation learning.

\section{Related Works}
\subsection{Instance Discrimination}
The concept of instance discrimination can be traced back to IR  \cite{DBLP:conf/cvpr/WuXYL18}, which conducts contrastive learning by treating each sample as a separate class, and pulling the positive samples closer while pushing the negative ones far away to learn instance-specific discriminative representation. 
MoCo \cite{DBLP:conf/cvpr/He0WXG20} adopts an online encoder and a momentum encoder to receive two views of a sample as the positive pair
. Additionally, a momentum updated queue is built to store negative samples.
SimCLR \cite{DBLP:conf/icml/ChenK0H20} adjusts the batch size as large as 4096 in the experiments to push SSL pre-training to the comparable effect as the supervised methods. Meanwhile, it also carefully sorts out the tricks that are very useful for improving the effect of SSL, such as longer training time, adding MLP projectors, or stronger data augmentation. Inspired by SimCLR \cite{DBLP:conf/icml/ChenK0H20}, He $et al.$ added the projector to MoCo \cite{DBLP:conf/cvpr/He0WXG20} and proposes MoCo-v2 \cite{DBLP:journals/corr/abs-2003-04297}, which refreshes SSL performance once again. Without adopting any negative samples, BYOL \cite{DBLP:conf/nips/GrillSATRBDPGAP20} adds a predictor to learn the map from the online encoder to the momentum encoder instead of displaying positive samples. Meanwhile, through the stop gradient mechanism, the negative samples are skillfully discarded. Simsiam \cite{DBLP:journals/corr/abs-2011-10566} lets the target encoder and the online encoder be the same and points out that the predictor and the stop gradient mechanism are sufficient conditions for training a strong SSL encoder for pre-training.

\subsection{Semantics Discovery of Instance}
\noindent {\bf Neighbors Discovery} AND \cite{DBLP:conf/icml/HuangDGZ19} discovers sample anchored neighborhoods to reason the underlying class decision boundaries. But it is restricted by the small size of local neighborhoods. PAD \cite{DBLP:conf/aaai/Huang0GZ20} bases on self-discovering semantically consistent groups of unlabelled
training samples with the same class concepts through a progressive affinity diffusion process. \cite{DBLP:conf/iccv/ZhuangZY19} trains an embedding function to maximize a metric of local aggregation, causing similar data instances to move together in the embedding space while allowing dissimilar instances to separate.

\noindent {\bf Deep Clustering} As a common technology for unlabeled data mining to learn higher-level visual understandings, deep clustering \cite{DBLP:conf/eccv/CaronBJD18,DBLP:conf/iclr/AsanoRV20a,DBLP:conf/eccv/NiuZWL20,DBLP:conf/cvpr/HuangGZ20,DBLP:conf/cvpr/ZhanX0OL20,DBLP:conf/iccv/JiVH19,DBLP:conf/nips/CaronMMGBJ20,zyf2} has been extended to learn in deep neural networks. DeepCluster \cite{DBLP:conf/eccv/CaronBJD18} is a representative method in alternate learning, which iteratively groups the features with k-means and uses the subsequent assignments to update the deep network.
Another recent work SeLa \cite{DBLP:conf/iclr/AsanoRV20a} makes cluster assignments by solving the optimal transport problem and alternatively performs representation learning and self-labeling. 
Another mode of deep clustering is pretext supervision, whose objective is to design a pretext task to use specific excuse objectives to learn label assignment and feature update simultaneously and indirectly impose the requirements for learning a good cluster. 
GATCluster \cite{DBLP:conf/eccv/NiuZWL20} designs four self-learning tasks with the constraints of transformation invariance, separability maximization, entropy analysis, and attention mapping to directly outputs semantic cluster labels without further post-processing.
PICA \cite{DBLP:conf/cvpr/HuangGZ20} learns the most semantically plausible data separation by maximizing the ``global” partition confidence of clustering solution. 
There is a lot of work turn to online clustering \cite{DBLP:conf/cvpr/ZhanX0OL20,DBLP:conf/iccv/JiVH19,DBLP:conf/nips/CaronMMGBJ20} to reduce error accumulation and the irrelevance of the pretext task to the downstream tasks occurring in offline clustering. ODC \cite{DBLP:conf/cvpr/ZhanX0OL20} designs and maintains two dynamic memory modules to perform clustering and network updating simultaneously. IIC \cite{DBLP:conf/iccv/JiVH19} maximizes the mutual information between the class assignments of each pair to output semantic labels. SwAV \cite{DBLP:conf/nips/CaronMMGBJ20} predicts the cluster assignment of a view from the representation of another view to simultaneously cluster the data while enforcing consistency between cluster assignments produced for different views of the same image, instead of comparing features directly as in traditional contrastive learning.

\subsection{Pixel Discrimination}
Global (instance-level) representations are efficient to compute but provide low-resolution features invariant to pixel-level variations. This might be sufficient for few tasks like image classification but are not enough for dense prediction tasks \cite{DBLP:conf/nips/PinheiroABGC20}. For better transference to downstream dense prediction tasks, \cite{DBLP:conf/nips/PinheiroABGC20,DBLP:journals/corr/abs-2011-10043,DBLP:journals/corr/abs-2011-09157} begin to focus on contrastive learning based on pixel discrimination. VaDeR \cite{DBLP:conf/nips/PinheiroABGC20} forces representations of pixels to be viewpoint agnostic through the correspondence of spatial coordinates, that is, the positive sample pairs come from the intersection of the two views, while the pixels from different images are negative samples. Pixpro \cite{DBLP:journals/corr/abs-2011-10043} adds a Pixel-to-Propagation
Module, emphasizing the consistency from pixel to propagation, which encourages spatially close pixels to be similar and can aid prediction in areas that belong to the same label. However, both these two methods don't consider the situation that the multiple views are not overlapping at all. DenseCL \cite{DBLP:journals/corr/abs-2011-09157} selects positive samples by ranking the similarities between all the pixels instead of utilizing spatial correspondence, which is no longer restricted by the requirement of the views' intersection.

\section{Methods}
\subsection{Sample Discrimination}
\noindent {\bf Instance Discrimination} For self-supervised representation learning, the breakthrough approaches are \cite{DBLP:conf/cvpr/WuXYL18,DBLP:conf/icml/ChenK0H20,DBLP:conf/cvpr/He0WXG20,DBLP:journals/corr/abs-2003-04297,DBLP:conf/nips/GrillSATRBDPGAP20,DBLP:journals/corr/abs-2011-10566}, which employ instance discrimination based on contrastive learning to learn good representations from unlabeled data. As our baseline is MoCo-v2 \cite{DBLP:journals/corr/abs-2003-04297}, we briefly introduce the instance-level contrastive learning  based on it. Two kinds of transformations $T_{a}$ and $T_{b}$ are randomly applied to a given sample $x_{i}$, and get two views $x_{i}^{a}=T_{a}(x_{i})$, $x_{i}^{b}=T_{b}(x_{i})$. $T_{a}$ and $T_{b}$ are from a set of transformations $T$. The details will be introduced in the experiments. After feeding them into an online encoder $f_{\mathit{\theta}}$ and a momentum encoder $f_{\hat{\mathit{\theta}}}$ respectively, we can get the feature vectors $h_{i}^{a} = f_{\mathit{\theta}}(x_{i}^{a})$ and $h_{i}^{b} = f_{\hat{\mathit{\theta}}}(x_{i}^{b})$ as shown in Figure~\ref{pipeline}. In order to project the feature vectors into an embedded space with a specific dimension, the two-layer global MLP projector is applied in MoCo-v2 \cite{DBLP:journals/corr/abs-2003-04297}, that is, $g_{a} (*)$ and $g_{b} (*)$ for different views. And the embeddings can be expressed as $z_{i}^{a}=g_{a}(h_{i}^{a})$ and $z_{i}^{b}=g_{b}(h_{i}^{b})$. The positive sample pairs are the two views from different transformations ($z_{i}^{a}$ and $z_{i}^{b}$), and the negative ones $z_{-}$ are from the momentum queue. A contrastive loss function InfoNCE \cite{DBLP:journals/corr/abs-1807-03748} is employed to pull the positive sample pairs closer while pushing them away from other negative keys:
\begin{equation}
   \mathcal{L}_{ins}=-log \frac{exp(s(z_{i}^{a}, z_{i}^{b}) /\tau_{ins} )}{exp(s(z_{i}^{a}, z_{i}^{b}) /\tau_{ins}) +\sum_{z_{-}}exp(s(z_{i}^{a}, z_{-})/\tau_{ins} ) },
\end{equation}
where $\tau_{ins}$ is a temperature to control the instance-level concentration degree of distribution \cite{DBLP:conf/cvpr/WuXYL18}. And the pair-wise similarity $s(a,b)$ is represented by cosine distance:
\begin{equation}
   s(a,b)=\frac{ab^{T}}{\left \| a \right \|\left \|b  \right \|}.
\end{equation}

\noindent {\bf Pixel Discrimination} The discrimination of instance-granularity regards each image as a separate individual, which will help to get a relatively global discriminative feature representation. Based on this, we conduct pixel discrimination in each image to get dense distinguished feature representation for each pixel. Similarly, we take the two views of a sample $x_{i}$ to feed into the two encoders. Different from the instance discrimination task, we replace the global projector of the previous instance-level with the dense projector, that is, $d_{a} (*)$ and $d_{b} (*)$. The corresponding dense feature representations are denoted as $v_{i}^{a}=d_{a}(h_{i}^{a})$ and $v_{i}^{b}=d_{b}(h_{i}^{b})$, and the contrastive learning is carried out in the dense embedding space:
\begin{equation}
   \mathcal{L}_{pix}=-log \frac{exp(s(v_{i}^{a}, v_{i}^{b}) /\tau_{pix} )}{exp(s(v_{i}^{a}, v_{i}^{b}) /\tau_{pix}) +\sum_{v_{-}}exp(s(v_{i}^{a}, v_{-})/\tau_{pix} ) },
\end{equation}
where $\tau_{pix}$ is a temperature to control the pixel-level concentration degree of distribution similarly. Following DenseCL \cite{DBLP:journals/corr/abs-2011-09157}, we get the correspondence of two dense views by calculating the distance of each pixel pair, then take the closest one as the positive pair:
\begin{equation}
   c_{i} = \mathop{\arg\max}_{j} s(v_{i}^{a},v_{j}^{b}),
\end{equation}
where $c_{i}$ is the most similar pixel index with the pixel vector $v_{i}^{a}$. Instead of adopting the spatial information to get the correspondence between two views \cite{DBLP:conf/nips/PinheiroABGC20,DBLP:journals/corr/abs-2011-10043}, the situation that the two views are not overlapped is considered.

\subsection{Semantics Mining}
After sample discrimination, there are two problems to be faced with. First, all the pixels are pushed away in each instance ($i.e.$ image), ignoring its inherent pixel-level semantic structure. Taking the image containing a dog and a cat in Figure~\ref{motivation} for example, all the pixels of the dogs should be closer to each other, and those of cats should also be closer to each other, while the pixels of these two objects should be pushed away.  Second, after the execution of pixel-level discrimination, there is still no connection between the pixels of different images. As shown in Figure~\ref{motivation}, the sample discrimination process does not consider that the pixels of the same kind of objects ($eg.$ dogs) in the two images should be drawn closer. On the contrary, it blindly pushed them farther instead, merely because they belong to different images. For the first problem, we propose to search for neighbors from multiple views,
and for the second problem, we try to conduct some cluster methods to reassign the label of each pixel among images ($i.e.$, cross images).

\begin{figure}
  \centering
  \includegraphics[width=0.85\linewidth]{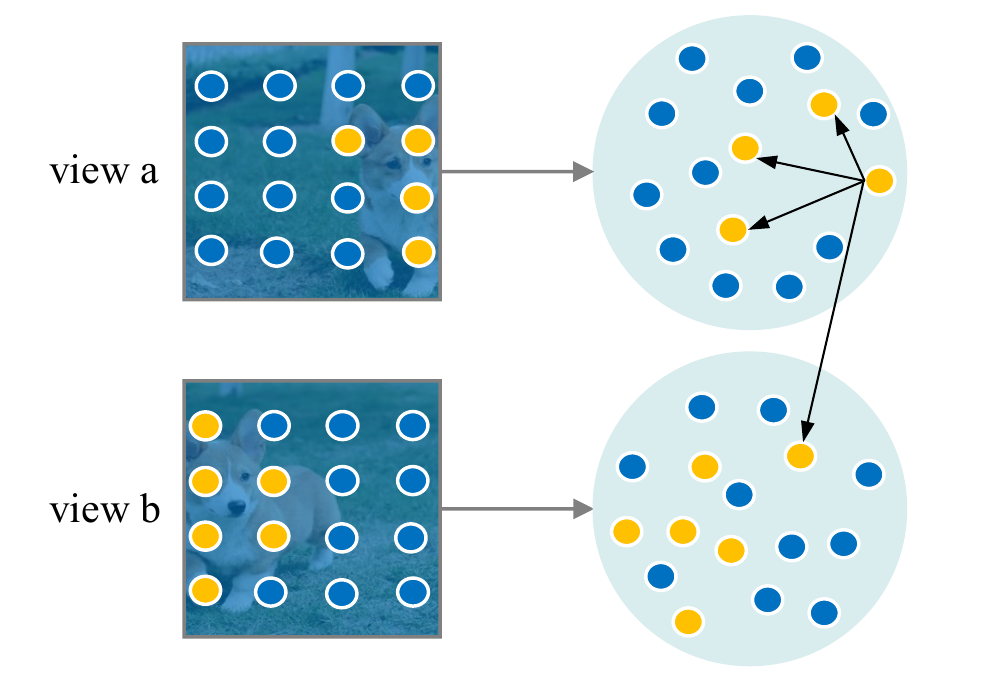}
  \caption{Illustration of neighbors discovery. The solid yellow circles are pixels from the object ``dog" in different views (a and b). }
  \label{fig3}
\end{figure}

\noindent {\bf Neighbors Discovery} Based on sample discrimination, we try to discover relative samples for each pixel to solve the first problem of lacking pixel-level semantic structures within images. We define these relative samples as the neighbors $\{n_{i}^{j}|j=\{1,..., N_{i}\}\}$ of the pixels, which should be treated as positive samples to draw closer. Note that $N_{i}$ is the number of neighbors for pixel $x_{i}$. The adjusted contrastive loss $\mathcal{L}_{nei}$ can be expressed as Eq.5. Through this additional pulled operation, the inherent pixel-level semantic structure within images is explored explicitly to a certain extent. Pixels are not only close to their augmented versions from different views but also adjacent with their neighbors from the same view, as shown in Figure~\ref{fig3}. For each pixel, the neighbors are discovered by ranking each pixel pair's similarity in one image. And we select the top-$N_{i}$ pixels as their neighbors. In fact, we can not only draw close to the neighbors in the same view, but also pull the neighbors from different views close. In theory, all these samples should belong to the same semantic category. In this paper, only the neighbors in the same view are discussed, and the performance is improved evidently, which benefits from the semantic information brought by the neighbors.

\begin{equation}
   \mathcal{L}_{nei}= -log \frac{exp(s(v_{i}^{a}, v_{i}^{b})/\tau_{pix})+ Nei}{exp(s(v_{i}^{a}, v_{i}^{b}) /\tau_{pix}) + Nei+\sum_{v_{-}}exp(s(v_{i}^{a}, v_{-})/\tau_{pix} ) }, \nonumber
\end{equation}
\begin{equation} 
   Nei = \sum_{N_{i}}exp(s(v_{i}^{a}, n_{i}^{j}) /\tau_{pix}).
\end{equation}

Furthermore, we try to construct a tripled relation of pixels ($i.e.$, a pixel $v_{i}^{a}$, its augmented version from another view $v_{i}^{b}$ and its j neighbors from the same view $n_{i}^{j}$). Our objective is to force the distance between $v_{i}^{a}$ and $v_{i}^{b}$ should be shorter than the distance between $v_{i}^{a}$ and its j-th neighbors $n_{i}^{j}$. The formal expression is
\setcounter{equation}{5}
\begin{equation}
   \mathcal{L}_{tri} = \sum_{j}^{N_{i}}\left [  s(v_{i}^{a},v_{i}^{b}) - s(v_{i}^{a},n_{i}^{j})+\alpha \right ]_{+},
\end{equation}
where $\alpha$ is the margin parameter and we take 0.3 in the experiments. The pair-wise similarity is defined as cosine distance in Eq.2.

\begin{figure}
  \centering
  \includegraphics[width=0.85\linewidth]{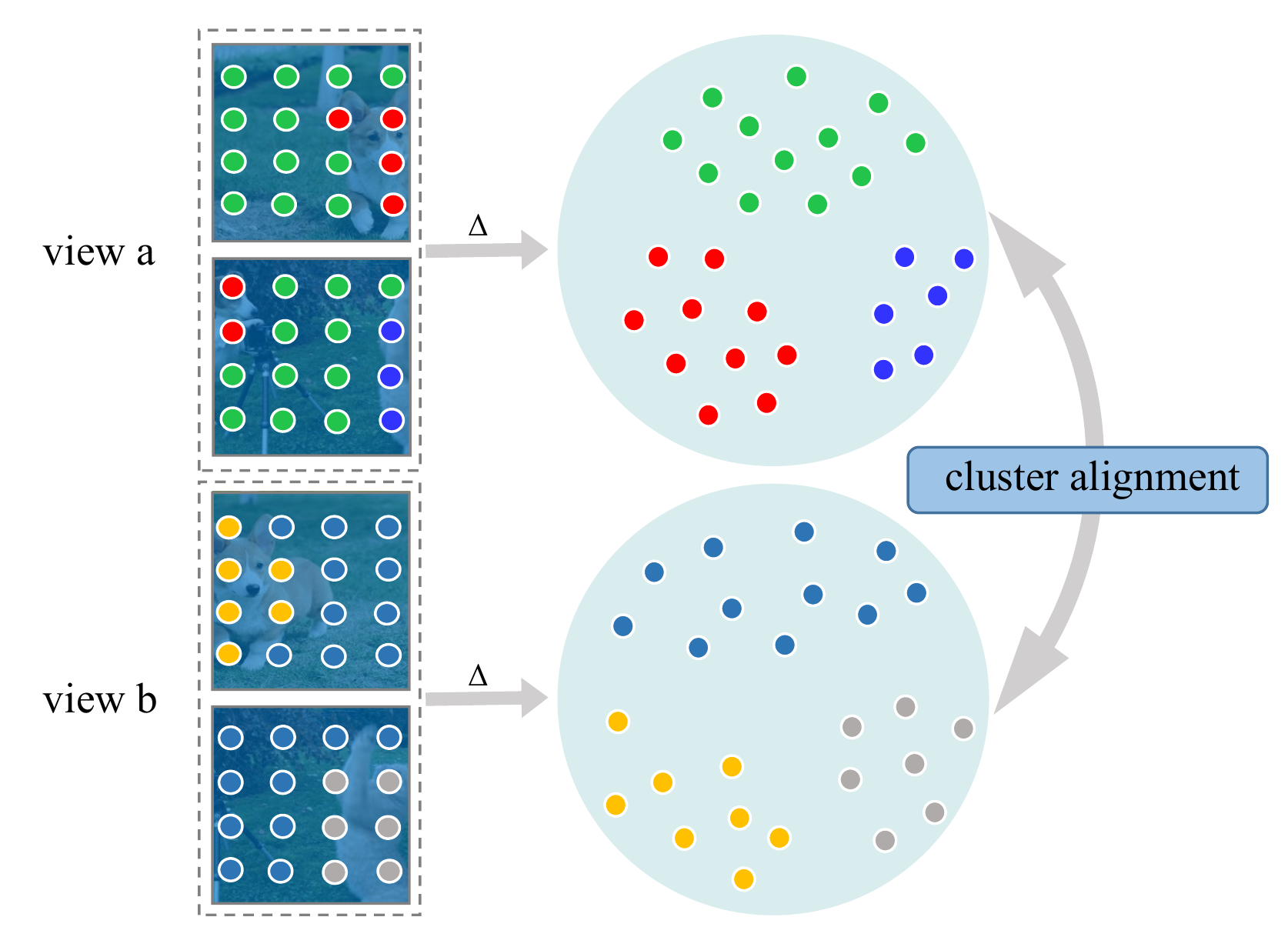}
  \caption{Illustration of cluster alignment. $\bigtriangleup$  denotes the clustering process, such as k-means or prototype mapping.}
  \label{fig4}
\end{figure}

\noindent {\bf Deep Clustering} For the second problem which lacks the global relations across the images, we explore some clustering methods to model the semantic category decision boundaries for high-level visual understandings. A natural practice is to perform k-means on the pixel-level embedded features in each image to get a certain number of clusters and then carry out contrastive learning for each cluster. The pixels within a specific cluster are closer to each other, and the samples among different clusters are pushed farther away. Theoretically, the clustering results in different images are various, and contrastive learning is required to carry out in each cluster and each image, which is very time-consuming and computation intensive. Therefore, it is necessary to explore a way to save time and effort simultaneously. Instead of contrasting the obtained features from the network offline, we skillfully design two modules ($i.e.$, KM and PM) to implement online clustering through cluster alignment as shown in Figure~\ref{fig4}.

{\bf DSC-KM} K-means is a general way to cluster samples but is time-consuming and only offline, which can not be used in large-scale application scenarios. We adjust the k-means algorithm to achieve online clustering. Specifically, all the pixels are clustered in a mini-batch, considering not only the pixels with similar semantics in one image, but also the pixels with similar semantics in different images. We compute the pixel centroid embeddings $e_{c}$ using the pixel embeddings only within a batch, and then contrast the $K$ centroid embeddings with Eq.7:
\begin{equation}
   \mathcal{L}_{KM}=-log \frac{exp(s(e_{c}^{a}, e_{c}^{b}) /\tau_{KM} )}{exp(s(e_{c}^{a}, e_{c}^{b}) /\tau_{KM}) +\sum_{e_{-}}exp(s(e_{c}^{a}, e_{-})/\tau_{KM} ) },
\end{equation}
where $\tau_{KM}$ is the temperature coefficient. We don't cluster the pixels within an image according to the general thinking, as it limits the semantic category relation modeling in an image,  which ignores the ``global” context of the training data, $i.e.$, the rich semantic relations between pixels across different images. DSC-KM can overcome the shortcomings of the previous pixel-level approaches without considering the semantic information and skillfully solve the problem of blindly pushing away all instances without considering the semantic relations among different instances. What's more, we realize online clustering to avoid time-consuming, which is one stroke in the third.

{\bf DSC-PM} We also explore a prototype-mapping approach to realize pixel clustering for efficiency. Specially,  we cluster the cross-image pixels and enforce the consistency between cluster assignments from different views of the same pixel simultaneously.  Given two pixel embeddings $v_{i}^{a}$ and $v_{i}^{b}$ from two views of $x_{i}$, we compute their codes (or cluster assignments) $q_{i}^{a}$ and $q_{i}^{b}$ by matching these embeddings to a set of $K$ prototypes $\{c_{1}, . . . , c_{K}\}$. We then set up a ``swapped” prediction problem \cite{DBLP:conf/nips/CaronMMGBJ20} with Eq.9.:
\begin{equation}
   \mathcal{L}_{PM} = \mathit{l}(v_{i}^{a},q_{i}^{b})+\mathit{l}(v_{i}^{b},q_{i}^{a}).
\end{equation}
The function $\mathit{l}(v,q)$ measures the matched degree between the feature $v$ and the code $q$, which can be represented by a cross-entropy loss between the code and the probability $p$ obtained by taking a softmax function.  The softmax function is made up of the dot products of $v_{i}^{a}$ and all prototypes in $C$:
\begin{equation}
   \mathit{l}(v,q)=\sum_{k}q^{k}\log{p^{k}} , ~ p^{k}=\frac{\exp{(v^{T}c^{k})}}{\sum_{k'}\exp{(v^{T}c^{k'}})}.
\end{equation}
Instead of contrasting the specific clustering results one by one, we force the alignment of cluster assignments from different views to make the network learn more semantic discriminative representation at the pixel level.

{\bf Discussion} DSC-KM and DSC-PM impose different constraints on the SSL model to explore the semantic decision boundary among pixels, both confirming this effectiveness obviously. DSC-KM requires the centroids of pixel clustering to be consistent under the two views, while DSC-PM forces each pixel's class assignment to be consistent. Comparatively, DSC-PM implements more strict constraints than DSC-KM to the model.

\subsection{Multi-granularity Framework}
Our proposed dense cross-image semantic contrastive learning framework considers multiple granularity representation learning to obtain not only low- and middle- but also high-level visual understandings shown in Figure~\ref{pipeline}. The overall objective function can be divided into three components according to different granularities in Eq.10, that is instance discrimination loss $\mathcal{L}_{ins}$, pixel discrimination loss $\mathcal{L}_{pix}$ and semantic discrimination loss $\mathcal{L}_{sem}$. Note that $\mathcal{L}_{sem}$ can be $\mathcal{L}_{nei}$, $\mathcal{L}_{tri}$, $\mathcal{L}_{KM}$ and $\mathcal{L}_{PM}$, only one of which will be used in our DSC model. 
Note that we have explored the influence of different orders of magnitude of weights on the model, and the experimental settings with the best performance are shown in the paper. 
\begin{equation}
 \mathcal{L}= \mathcal{L}_{ins} +\mathcal{L}_{pix} + \mathcal{L}_{sem}.
\end{equation}

\section{Experiments}
Following recent self-supervised methods \cite{DBLP:journals/corr/abs-2003-04297,DBLP:journals/corr/abs-2011-09157}, we adopt ResNet-50 \cite{DBLP:conf/cvpr/HeZRS16} as our backbone. The model is pre-trained on ImageNet and MS COCO respectively. Our objective is to verify the transferred ability of the feature representation learned by the pre-trained model to the downstream dense prediction tasks. Therefore, we evaluate the object detection and segmentation tasks on various datasets. Specifically, it's object detection on PASCAL VOC, semantic segmentation on PASCAL VOC and Cityscapes, object detection and instance segmentation on MS COCO. Besides, we conduct our ablation study with object detection and semantic segmentation on PASCAL VOC, adopting the model pre-trained on MS COCO.

In the pretraining stage, we follow the data augmentation of DenseCL \cite{DBLP:journals/corr/abs-2011-09157}, which contains two randomly sampled crops from the image and resized to 224 × 224 with a random horizontal flip, followed by a Random Grayscale. At the same time, ColorJitter and GaussianBlur are randomly selected.

\begin{table}[htbp]
  \centering
  \caption{The performance of PASCAL VOC object detection. CC and IN indicate the pre-training models trained on MS COCO and ImageNet respectively. $*$ represents our re-implementation.}
  \setlength{\tabcolsep}{5mm}
    \begin{tabular}{l|ccc}
    \toprule
    pre-train & \multicolumn{1}{c}{$AP$} & \multicolumn{1}{c}{$AP_{50}$} & \multicolumn{1}{c}{$AP_{75}$} \\
    \hline
     MoCo-v2 CC* & 52.1  & 79.0    & 56.7 \\
    DenseCL CC* & 56.4  & 81.8  & 62.7 \\
    DSC-KM (Ours) & 57.0&82.1&63.0 \\
    DSC-PM (Ours)  & \textbf{57.2} & \textbf{82.3} & \textbf{63.4} \\
    \hline
    SimCLR IN \cite{DBLP:conf/icml/ChenK0H20}& 51.5  & 79.4  & 55.6 \\
    BYOL IN \cite{DBLP:conf/nips/GrillSATRBDPGAP20} & 51.9  & 81.0    & 56.5 \\
    MoCo IN \cite{DBLP:conf/cvpr/He0WXG20} & 55.9  & 81.5  & 62.6 \\
    Moco-v2 IN* & 57.1  &  82.0 & 63.9\\
    DenseCL IN* &  58.4  & 82.7  & \textbf{65.7} \\
    DSC-KM (Ours)  &  \textbf{58.7} & 82.7 &65.6  \\
    DSC-PM (Ours)  & 58.6 & \textbf{82.8} & 65.6 \\
    \bottomrule
    \end{tabular}%
  \label{table1}%
\end{table}%

\subsection{Experimental Settings}
\noindent {\bf Pre-training} Followed MoCo-v2 \cite{DBLP:journals/corr/abs-2003-04297}, we utilize the same training settings, like  $\tau$ is 0.2 (in our work, $\tau_{ins}$, $\tau_{pix}$, and $\tau_{cen}$ are all set to 0.2). The initial learning rate is 0.3. A SGD optimizer is adopted in the models with a Nesterov momentum of 0.9 and a weight decay of 1e-4. All the models are optimized on 8 V100 GPUs with a cosine learning rate decay schedule and a mini-batch size of 256. We train 800 epochs for MS COCO and 200 epochs for ImageNet.  Note that we re-implement DenseCL and MoCo-v2, and can achieve comparable results. For a fair comparison, all the experiments share the same training settings including other re-implemented methods and ours, and we use the re-implemented results to represent their performance as we can get comparable results with their papers.

\noindent {\bf Downstream tasks training} In order to further evaluate the quality of feature representation learned by pre-trained models, we apply the models to various downstream dense prediction tasks followed by fine-tuning. For PASCAL VOC object detection, we fine-tune a Faster R-CNN \cite{DBLP:journals/pami/RenHG017} detector with C4-backbone adopting a standard 2x schedule with \cite{wu2019detectron2}. The training set is $train2007\&2012$, and the test set is $test2007$. For PASCAL VOC and Cityscapes semantic segmentation, we fine-tune a FCN \cite{DBLP:journals/pami/ShelhamerLD17} model with 20k iterations and 40k iterations respectively. Different from DenseCL \cite{DBLP:journals/corr/abs-2011-09157}, we adopt $train2012$ for training and $val2007$ for evaluating on PASCAL VOC, while $train\_aug2012$ was used for training in the former. And for Cityscapes, $leftImg8bit$ is used for training and evaluating. For object detection and instance segmentation on MS COCO, we fine-tune a Mask R-CNN \cite{DBLP:journals/pami/HeGDG20} detector with a FPN model as the backbone, adopting a standard 1x schedule. The model is trained on $train2017$ and evaluated on $val2017$. We uses $AP$, $AP_{50}$ and $AP_{75}$ to evaluate for object detection, $AP^{m}$, $AP_{50}^{m}$ and $AP_{75}^{m}$ for instance segmentation, and  mean Intersection over Union ($mIoU$) for semantic segmentation.

\subsection{Main Results}
\noindent {\bf PASCAL VOC object detection} We compare our methods with state-of-the-art approaches on PASCAL VOC object detection. As shown in Table~\ref{table1}, for the model pre-trained on MS COCO and ImageNet, both DSC-KM and DSC-PM can get better performance than another two previous methods. Especially, DSC-PM achieves 5.1\% $AP$ higher performance than MoCo-v2 and 0.8\% $AP$ higher than the baseline DenseCL on MS COCO pre-trained models, DSC-KM is 1.6\% $AP$ higher than MoCo-v2 and 0.3\% $AP$ higher than DenseCL on ImageNet pre-trained models. The promotion shown in Table~\ref{table1} indicates that the gap between the pre-trained models and downstream dense prediction tasks will narrow substantially with the help of reasoning semantic category decision boundaries.

\begin{table}
\caption{The performance of semantic segmentation on (a) PASCAL VOC and (b) Cityscapes. CC and IN indicate the pre-training models trained on MS COCO and ImageNet respectively. $*$ represents our re-implementation.}
    \begin{minipage}{0.48\linewidth}
      \centering
    (a)~PASCAL VOC
    
      \begin{tabular}{l|c}
       \toprule
    pre-train & \multicolumn{1}{c}{$mIoU$} \\
    \hline
    MoCo-v2 CC* & 48.6 \\
    DenseCL CC* & 56.7 \\
    DSC-KM (Ours) & 57.7 \\
    DSC-PM (Ours) &\textbf{57.9}\\
    \hline
    Moco-v2 IN* & 56.3  \\
    DenseCL IN* & 58.9 \\
    DSC-KM (Ours)  & 59.3 \\
    DSC-PM (Ours) & \textbf{59.6}\\
    \bottomrule
    
      \end{tabular}
      
    \end{minipage}
    \begin{minipage}{0.48\linewidth}
      \centering
    (b)~Cityscapes
    
      \begin{tabular}{l|c}
        \toprule
    pre-train & \multicolumn{1}{c}{$mIoU$} \\
    \hline
     MoCo-v2 CC* & 72.4 \\
    DenseCL CC* & 75.3 \\
    DSC-KM (Ours) & \textbf{75.7}\\
    DSC-PM (Ours) & 75.5  \\
    \hline
    Moco-v2 IN* &  72.6 \\
    DenseCL IN* & 75.5 \\
    DSC-KM (Ours) & \textbf{76.0} \\
    DSC-PM (Ours) & 75.6 \\
    \bottomrule
    
      \end{tabular}
      
    \end{minipage}
    \label{table3}
  \end{table}

\noindent {\bf PASCAL VOC and Cityscapes semantic segmentation} Table~\ref{table3} (a) demonstrates PASCAL VOC semantic segmentation pre-trained on MS COCO and ImageNet respectively. Both DSC-KM and DSC-PM significantly improve the performance with a large margin. Especially,  DSC-PM outperforms MoCo-v2 with 9.3\% $mIoU$ and DenseCL with 1.2\% $mIoU$ when pre-trained on MS COCO, and outperforms MoCo-v2 with 3.3\% $mIoU$ and DenseCL with 0.7\% $mIoU$ when pre-trained on ImageNet, which verifies the effectiveness of the semantic category decision boundary modeling again. In Table~\ref{table3} (b), we report the semantic segmentation results on Cityscapes. For the models pre-trained on MS COCO, DSC-KM is 3.3\% higher than MoCo-v2 and 0.4\% higher than DenseCL. And for the models pre-trained on ImageNet, DSC-KM is 3.4\% higher than MoCo-v2 and 0.5\% higher than DenseCL. The significant performance improvements in semantic segmentation tasks demonstrate that the semantic category labels assignment in our methods is more accurate than any other SSL pre-training approach.

\begin{table}[htbp]
  \centering
  \caption{The performance of MS COCO object detection and instance segmentation. CC and IN indicate the pre-training models trained on MS COCO and ImageNet respectively. $*$ represents our re-implementation. }
  \setlength{\tabcolsep}{1.8mm}
    \begin{tabular}{l|ccc|ccc}
    \toprule
    pre-train & \multicolumn{1}{c}{$AP^{b}$} & \multicolumn{1}{c}{$AP_{50}^{b}$} & \multicolumn{1}{c|}{$AP_{75}^{b}$} & \multicolumn{1}{c}{$AP^{m}$} & \multicolumn{1}{c}{$AP_{50}^{m}$} & \multicolumn{1}{c}{$AP_{75}^{m}$} \\
    \hline
    MoCo-v2 CC* & 37.0  & 55.9  & 40.2  & 33.5  & 53.1  & 35.9 \\
    DenseCL CC* & 38.8  & 58.4  & 42.6  & 35.1  & 55.4  & 37.7 \\
    DSC-KM (Ours)  & \textbf{39.2} & \textbf{58.8} & \textbf{42.8}& \textbf{35.5} & \textbf{55.9} & \textbf{38.0}\\
    DSC-PM (Ours) & 39.0 & 58.6&42.5&35.1&55.5&37.7 \\
    \hline
    Moco-v2 IN* &  38.9     & 58.5  &42.5  &35.2& 55.6 & 37.8 \\
    DenseCL IN* &  39.2& 58.7 & 42.9 &35.5 &56.0 & 37.7 \\
    DSC-KM (Ours)  & 39.4 &58.8  &43.0  & 35.6  &56.1 &38.1  \\
    DSC-PM (Ours)  &  \textbf{39.4} & \textbf{58.9} &\textbf{43.2}  & \textbf{35.7}  & \textbf{56.1} & \textbf{38.3} \\
    \bottomrule
    \end{tabular}%
  \label{table2}%
\end{table}%

\noindent {\bf MS COCO object detection and instance segmentation} The results of MS COCO object detection and instance discrimination are shown in Table~\ref{table2}. With MS COCO pre-training, DSC-KM is 2.2\% $AP^{b}$,  2.0\%  $AP^{m}$ higher than MoCo-v2 and 0.4\% $AP^{b}$, 0.4\%  $AP^{m}$ higher than DenseCL. And for ImageNet, DSC-PM achieves 0.5\% $AP^{b}$, 0.5\% $AP^{m}$ higher performance than MoCo-v2 and 0.2\% $AP^{b}$, 0.2\% $AP^{m}$ higher performance than DenseCL. The improvements are limited both in MS COCO and ImageNet pre-trained models as MS COCO contains of a lot of authentic scenario images, which is still challenging for SSL pre-training to cope with difficulties in visual tasks of complex scenes.

\begin{table}[htbp]
  \centering
  \caption{Comparison of different semantic strategies for PASCAL VOC object detection and semantic segmentation. ``-" represents our baseline without any semantic strategies.}
  \setlength{\tabcolsep}{4.2mm}
    \begin{tabular}{l|ccc|c}
    \toprule
    strategy & \multicolumn{1}{c}{$AP$} & \multicolumn{1}{c}{$AP_{50}$} & \multicolumn{1}{c|}{$AP_{75}$} & \multicolumn{1}{c}{$mIoU$} \\
    \hline
    - & 56.4  & 81.8    & 62.7  &56.7  \\
    Neighbor & 56.6  & 81.6  & 63.0  &57.5    \\
    Triplet & 55.5  & 80.9  & 61.4 & 53.5 \\
    CE    & 56.8  & 81.9  & 63.0   & \textbf{58.1} \\
    KM & 56.8  & 81.9  & 62.8 & 57.7   \\
    PM  & \textbf{57.1}  & \textbf{82.2}  & \textbf{63.3} &57.9   \\
    \bottomrule
    \end{tabular}%
  \label{table4}%
\end{table}%
\vspace{-3pt}
\subsection{Ablation Study}
We conduct abundant ablation studies to explore the importance of each component in our framework. Specially, we first compare various strategies to supply the semantic category information. Moreover, we discuss the influence of the number of clusters on our performance. Besides, we also study the significance of discriminative representation learning in different granularities.

\begin{figure}
  \centering
   \includegraphics[width=\linewidth]{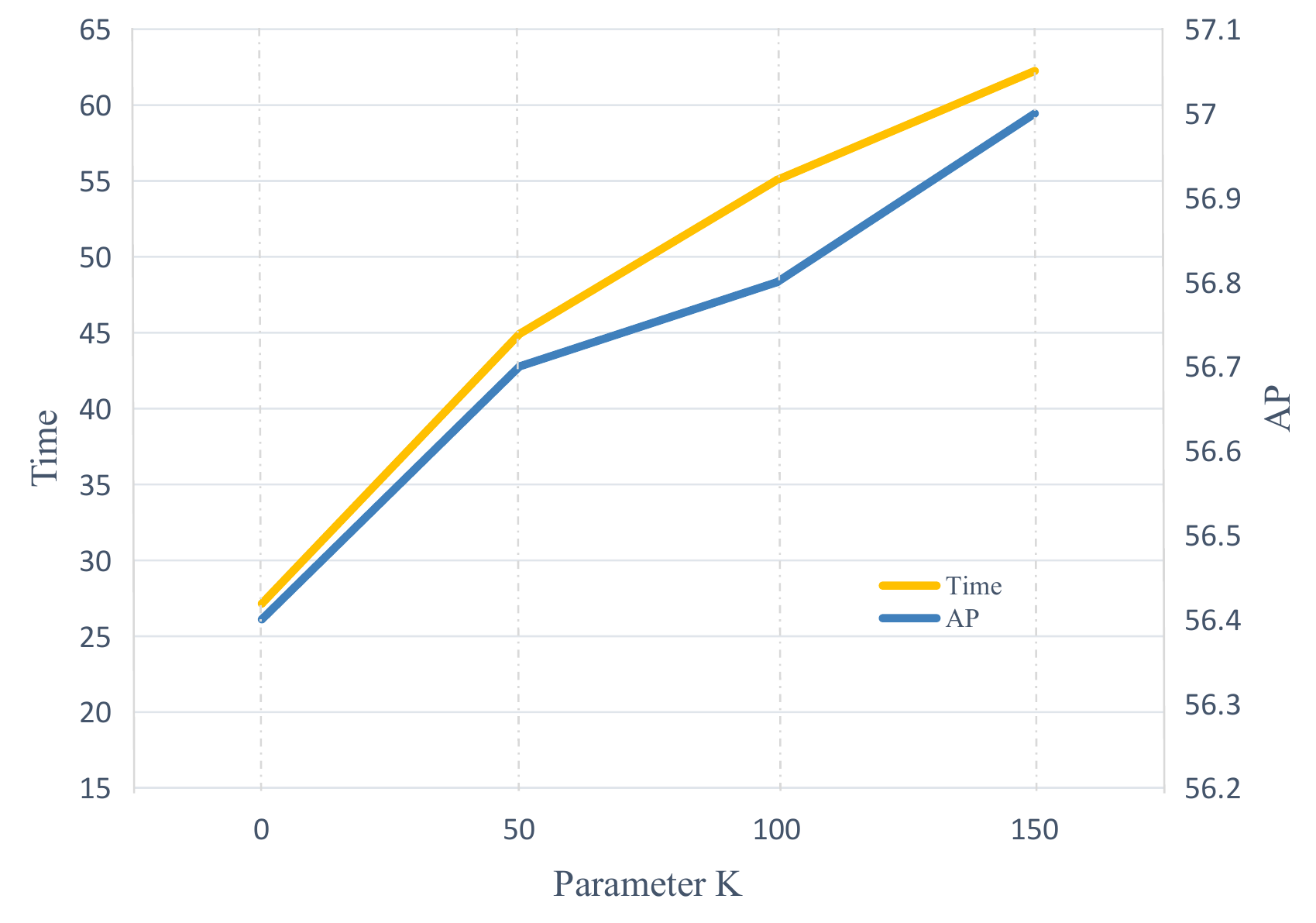}

  \caption{The performance and efficiency with different $K$ for DSC-KM on PASCAL VOC object detection.}
  \vspace{-10pt}
  \label{fig5}
\end{figure}

\noindent {\bf Semantic strategies} We explore five strategies to model the semantic category decision boundaries explicitly in Table~\ref{table4}.  ``Neighbor" means the neighbor-discovery method we mentioned before, and we take $N_{i}$ as 1. ``Triplet" constructs a triple with one pixel, its augmented version, and its neighbor in the same view, performing triplet loss based on pixel discrimination. ``CE" adopts cross-entropy loss for the cluster codes obtained by prototype mapping. ``PM" and ``KM" denote the semantics-mining methods we mentioned before. The experimental results show that all the strategies improve the performance of the downstream dense prediction tasks to some extent, indicating that the supplement of semantics for pixels can help get better semantic structures in the dataset. Moreover, we can see that the performance of ``CE", ``PM" and ``KM" is better than `Neighbor" and ``Triplet", which demonstrates that exploring global relations among cross-images pixels is more effective than mining the local relations of the pixels within a single image.  

\noindent {\bf Number of clusters K} We explore the effect of the number to cluster the pixel embeddings from the backbone. Figure~\ref{fig5} shows that in DSC-KM, with the increase of $K$, downstream tasks' performance becomes better, indicating that moderately over clustering is more beneficial for semantic representation learning. While the time-consuming situation also becomes serious along with the growth of $K$. Balancing the performance improvement and the time cost, we choose $K=100$ in our experiments for KM and $K=150$ for PM.


\noindent {\bf Granularities of representation learning} The influence of each component corresponding to different visual granularities is investigated, as shown in Table~\ref{table5}. With the increase of granularity considering in the framework, the performance of downstream tasks shows a progressive upward trend. By jointly learning representation in multiple granularities, the DSC model gets not only low- and middle-level visual understandings on an individual instance or pixel level, but also obtains high-level visual understandings on a semantic category level. This kind of multi-granularity consideration is beneficial for an accurate category assignment when carrying out downstream dense prediction tasks. 
We have carried out experiments with only pixel or semantic granularity, but both models can't converge. 
It is also mentioned in DenseCL that contrastive learning only at pixel-level can't converge.

\begin{figure}[ht]
  \centering
  \includegraphics[width=\linewidth]{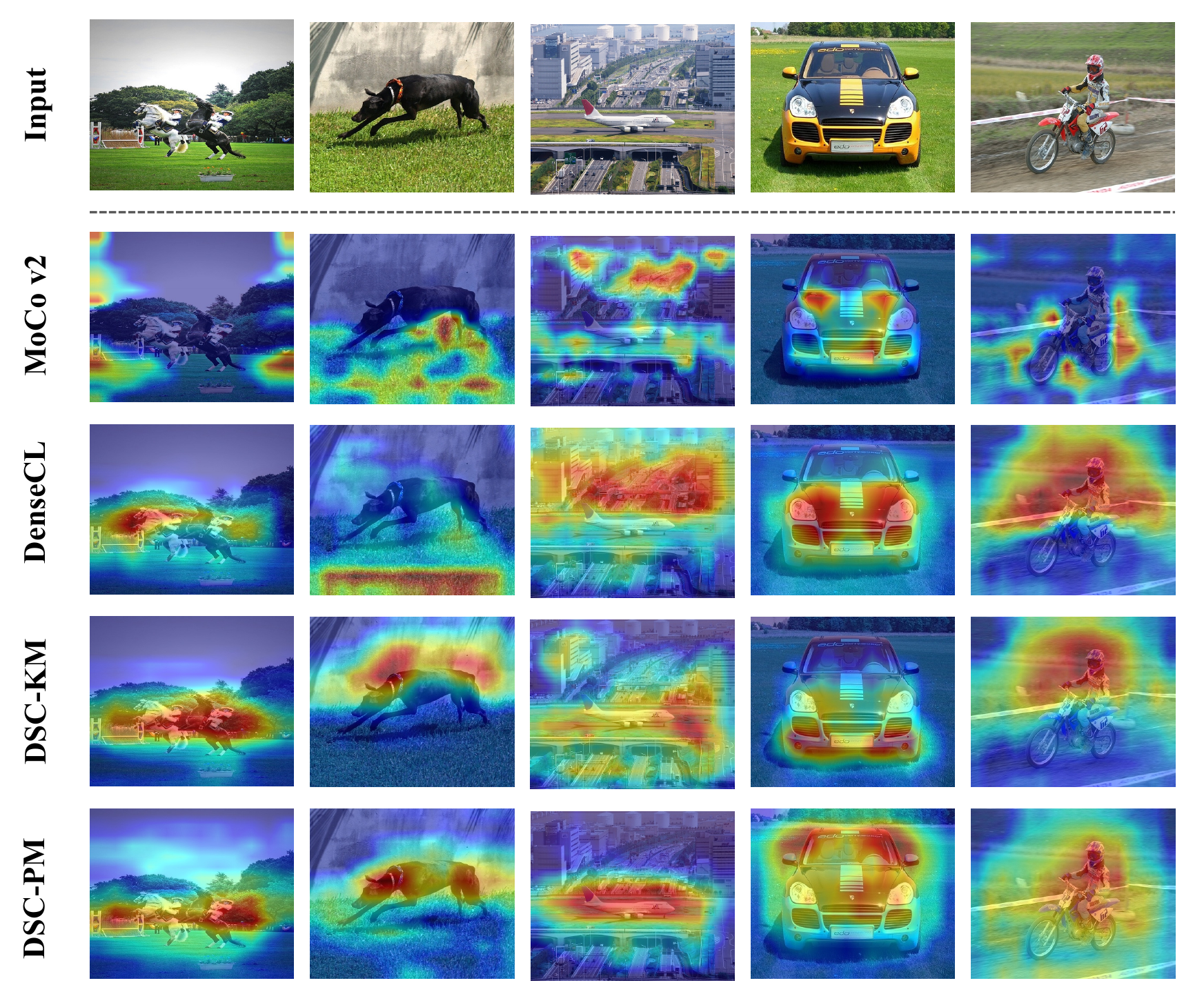}

  \caption{Comparison of the heat map visualization under MoCo-v2, DenseCL, our DSC-KM and DSC-PM by Grad-CAM \cite{DBLP:conf/iccv/SelvarajuCDVPB17} on MS COCO dataset.}
  \vspace{-5pt}
  \label{vis_heatmap}
\end{figure}
\begin{table}[htbp]
  \centering
  \caption{The influence of different granularities on PASCAL VOC object detection and semantic segmentation.}
   \setlength{\tabcolsep}{1.8mm}
    \begin{tabular}{ccc|ccc|c}
    \toprule
    instance & pixel & semantics & $AP$    & \multicolumn{1}{c}{$AP_{50}$} & \multicolumn{1}{c|}{$AP_{75}$} & \multicolumn{1}{c}{$mIoU$}  \\
    \hline
    \checkmark     &      &      &  54.7     &  81.0     &   60.6  &48.6  \\
    \checkmark    & \checkmark     &      & 56.4 & 81.8& 62.7 & 56.7\\
     \checkmark   & \checkmark     & \checkmark     & \textbf{57.1}  & \textbf{82.2}  & \textbf{63.3} & \textbf{57.9}   \\
    \bottomrule
    \end{tabular}%
    \vspace{-3pt}
  \label{table5}%
\end{table}%

\subsection{Visualization}
\noindent{\bf Feature representation visualization} More explicit boundary information of the objects is required in the downstream dense prediction tasks. To further understand the feature representation learned by different pre-trained models, we utilize Grad-CAM \cite{DBLP:conf/iccv/SelvarajuCDVPB17} to visualize the heat maps of these models pre-trained on MS COCO. As illustrated in Figure~\ref{vis_heatmap}, MoCo-v2 is more interested in distinguishing features conducive to classification, while pixel-level methods (DenseCL and ours) focus on regional features for downstream dense prediction tasks. 
Especially, our DCS model is more sensitive to object boundary information than DenseCL as we focus on mining the inherent semantic relations among pixels.

\begin{figure}[ht]
  \centering
  \includegraphics[width=\linewidth]{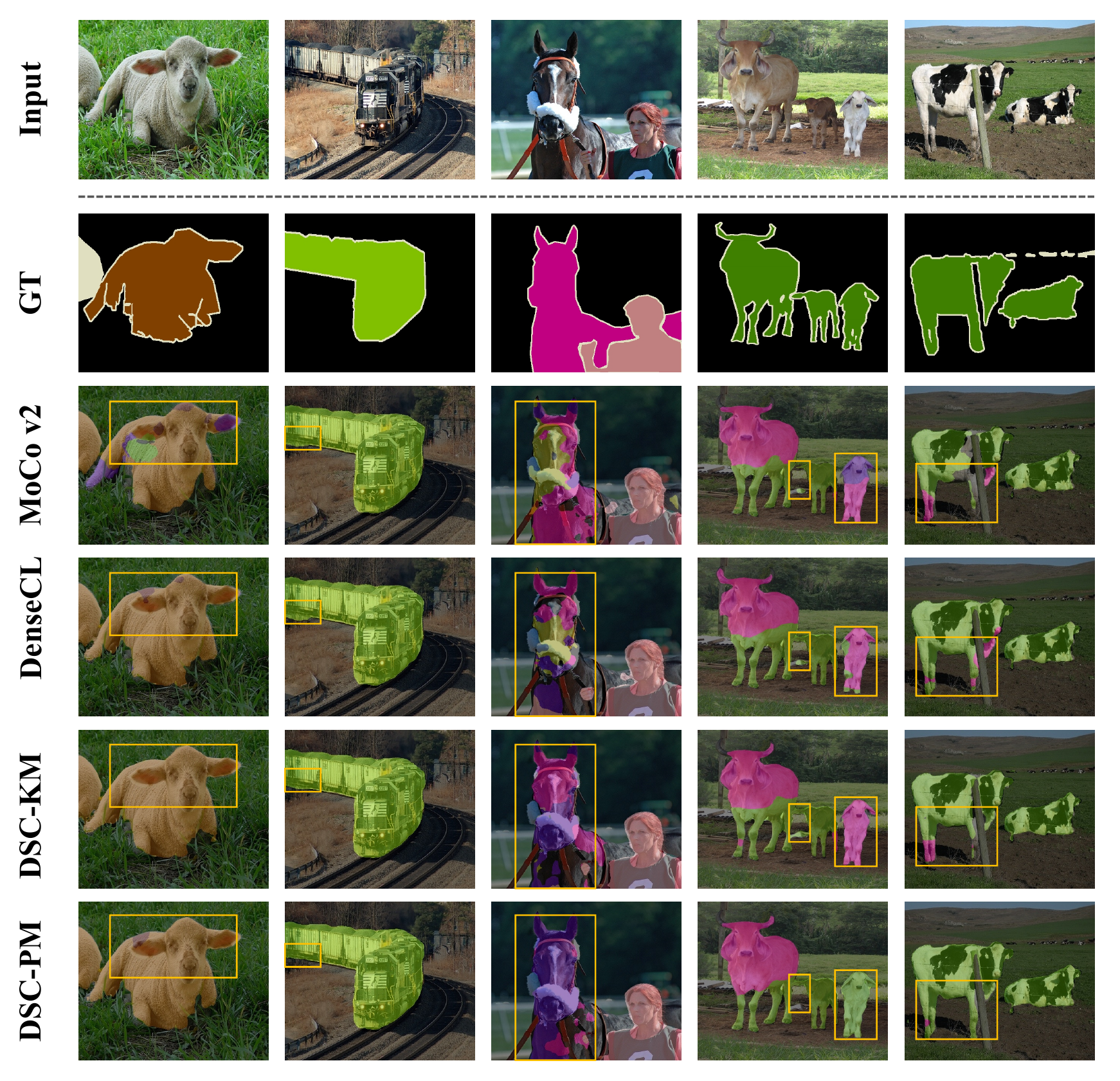}

  \caption{Comparison of semantic segmentation under MoCo-v2, DenseCL, our DSC-KM and DSC-PM on PASCAL VOC dataset.}
  \vspace{-5pt}
  \label{vis_seg}
\end{figure}

\noindent{\bf Downstream task visualization} We visualize the transferability of different models pre-trained on MS COCO fine-tuning on PASCAL VOC semantic segmentation task. From the third row of  Figure~\ref{vis_seg}, we can find that MoCo-v2 is more prone to misjudge the pixels' category as it only focuses on discriminative learning in the instance level, which lacks denser observation in feature representations. Despite that DenseCL is devoted to pixel discrimination task, the fourth row of Figure~\ref{vis_seg} shows that there are still some mistakes in pixel category assignment. It has a lot to do with the fact that DenseCL doesn't explore the semantic relations among pixels. With our DSC model, the performance becomes better in the segmentation task, proving the effectiveness of our semantics-mining methods for downstream dense prediction tasks once again. 

\section{Conclusion}
In this work, for the first time we have proposed Dense Semantic Contrast (DSC) for self-supervised visual representation learning, which models semantic category decision boundary at the pixel level to meet the command of semantic representation in downstream dense prediction tasks. What's more, a dense cross-image semantic contrastive framework has been constructed for multi-granularity pre-training representation learning, considering low-, middle- and high-level visual understandings simultaneously. The experimental results indicate that the effectiveness of making up for the absence of semantic category relation in our methods. We have skillfully designed an approach to align the semantic category assignment, forcing the network to learn more semantic discriminative feature representation implicitly. In the future, we will explore more skillful means to mine the semantic relationships among the datasets. What's more, we will consider hierarchical idea \cite{lxn1} for clustering.  We expect that the first exploration of modeling semantic category decision boundaries may inspire more related works in pixel semantic mining.
\begin{acks}
Supported by the Open Research Project of the State Key Laboratory of Media Convergence and Communication, Communication University of China, China (No. SKLMCC2020KF004), the Beijing Municipal Science \& Technology Commission (Z191100007119002), the Key Research Program of Frontier Sciences, CAS, Grant NO ZDBS-LY-7024, the National Natural Science Foundation of China (No. 62006221), and CAAI-Huawei MindSpore Open Fund.

\end{acks}

\bibliographystyle{ACM-Reference-Format}
\balance
\bibliography{sample-base}










\end{document}